\documentclass[sigconf]{acmart}
\AtBeginDocument{%
  }

\copyrightyear{2025} 
\acmYear{2025} 
\setcopyright{cc}
\setcctype{by}
\acmConference[FAccT '25]{The 2025 ACM Conference on Fairness, Accountability, and Transparency}{June 23--26, 2025}{Athens, Greece}
\acmBooktitle{The 2025 ACM Conference on Fairness, Accountability, and Transparency (FAccT '25), June 23--26, 2025, Athens, Greece}\acmDOI{10.1145/3715275.3732091}
\acmISBN{979-8-4007-1482-5/2025/06}
\begin{document}

\title{A Trustworthiness-based Metaphysics of Artificial Intelligence Systems}

\author{Andrea Ferrario}
\orcid{0000-0001-9968-9474}
\affiliation{%
  \institution{Institute of Biomedical Ethics and History of Medicine, University of Z\"urich}\city{Z\"urich}  \country{Switzerland}
  }
  \affiliation{%
  \institution{ETH Z\"urich}
  \city{Z\"urich}
  \country{Switzerland}
}
\email{aferrario@ethz.ch}

\renewcommand{\shortauthors}{Ferrario A.}

\begin{abstract}
Modern AI systems are man-made objects that leverage machine learning to support our lives across a myriad of contexts and applications. Despite extensive epistemological and ethical debates, their metaphysical foundations remain relatively under explored. The orthodox view simply suggests that AI systems, as artifacts, lack well-posed identity and persistence  conditions---their metaphysical kinds are no \emph{real} kinds. In this work, we challenge this perspective by introducing a theory of metaphysical identity of AI systems. We  do so by characterizing their kinds and introducing identity criteria---formal rules that answer the questions ``When are two AI systems the same?'' and ``When does an AI system persist, despite change?'' Building on Carrara and Vermaas' account of fine-grained artifact kinds, we argue that AI trustworthiness provides a lens to understand AI system kinds and formalize the identity of these artifacts by relating their functional requirements to their physical make-ups. 
The identity criteria of AI systems are determined by their trustworthiness profiles---the collection of capabilities that the systems must uphold over time  throughout their artifact histories, and their effectiveness in maintaining these capabilities.
Our approach suggests that the identity and persistence of AI systems is sensitive to the socio-technical context of their design and utilization via their trustworthiness, providing a solid metaphysical foundation to the epistemological, ethical, and legal discussions about these artifacts.
\end{abstract}



\maketitle

\section{Introduction}
In everyday life, we count and classify seashells on the beach and stars in the night sky. We individuate mushrooms and birds while walking in the woods. There is something about natural entities such as seashells, stars, mushrooms, and birds that enables us to identify and categorize them in different metaphysical classes---referred to as `kinds' \citep{lowe2015more}---based on  principles that govern their nature \citep{wiggins2001sameness}. Importantly, we can compare  individual entities within these kinds and determine whether they are identical by applying metaphysically-grounded rules known as \emph{identity criteria} \citep{lowe1983identity,lowe2014ontology,wiggins2001sameness}. These criteria also allow us discussing the different types of change these entities can undergo and assess whether these changes disrupt their existing over time, i.e., their persistence. This \textbf{theory of metaphysical identity} formalizes the identity and persistence of seashells, stars, mushrooms, and birds, granting them `metaphysical respectability' \citep{lowe2015more}. In fact, as Quine famously stated: ``no entity without identity'' \citep[pag.~23]{quine1969}. Furthermore, \emph{knowing about} these classes, identity criteria, and persistence conditions is important for our lives, allowing us to distinguish between bivalves and gastropods, black holes and supernovas, edible and poisonous mushrooms, as well as preserving endangered bird species: metaphysics guides ethics.

For a long time, philosophers have debated whether artifacts---objects humans design to fulfill specific functions---enjoy a metaphysics comparable to those of natural entities and living beings. In fact, discussing the theory of metaphysical identity of statues, clocks, smartphones, and freight trains rapidly gives rise to philosophical puzzles due to their distinctive properties. These include  their capacity to undergo  different types of transformations, such as disassembly and reassembly over time. The Ship of Theseus puzzle  is the most famous illustration of this problem \citep{hobbes1655dec}. Consequently, a widely held view is that artifacts are metaphysically deficient entities---their kinds are not \emph{real} kinds \citep{wiggins1980samething,carrara2009fine}, lacking well-posed identity criteria and persistence conditions \citep{lowe2014real}. This means that  
we cannot precisely individuate and count spoons, chairs, and ships or meaningfully define their persistence as material objects subjected to change over time.

In this work, we challenge this perspective on artifact kinds by introducing a \textbf{theory of metaphysical identity of modern artificial intelligence (AI) systems}. Namely, we argue that AI system kinds are real kinds: these artifacts have well-posed identity criteria and persistence conditions. 

Our approach proceeds in three steps. First, we conceptualize modern AI systems---referred to here simply as `AI systems'---as artifacts designed to employ machine learning models to assist humans across virtually all domains of society. Although these systems manifest in diverse shapes and forms they  share a fundamental trait: the necessity of retraining their machine learning models, a process that changes their logical and physical make-ups and that we posit as intrinsic to their nature as artifacts. Second, we apply Carrara and Vermaas' \emph{function}+ framework for the realism of artifact kinds \citep{carrara2009fine} to the case of AI systems. To do so, we (1) characterize AI system kinds through their `techno-function'---the technical capability they are designed to fulfill, e.g., `to predict individualized credit risk scores,' and (2) identify the \textbf{trustworthiness of AI} as their `operational principle' \citep{polanyi2012personal}. Here, the trustworthiness of an AI system
is a normatively rich and composite property that encompasses the capabilities an AI system must maintain to function correctly within its context of application
\citep{Floridi2019,Mattioli2024,Li2023,ferrario2024justifying,jacovi2021formalizing}. 
Finally, our identity criterion for AI systems within a given kind stipulates that two AI systems are identical \emph{as systems of that kind} if and only if they execute the same techno-function and adhere to the same trustworthiness criteria, while keeping compatible levels of trustworthiness. We project this criterion into synchronic, referring to identity at a given time, and diachronic, referring to identity at different times, perspectives. As a result, our identity theory allows for the identification of AI systems with differing physical compositions, provided they satisfy compatible functional and operational requirements that are encoded in their trustworthiness. Hence, our theory introduces a trustworthiness-based conception of `technological identity’ for artifacts, in line with Carrara and Vermaas' framework \citep{carrara2009fine}. While consistent with Lowe's view of complex artifacts as real kinds \citep{lowe2014real}, our approach  emphasizes that \textbf{the identity criteria and persistence conditions of AI systems are contingent upon the socio-technical environments that shape their design and dictate their correct functioning}.

\vspace{0.25cm}

Our contribution tackles an open point in the philosophy of modern AI. While current research predominantly focuses on epistemological and ethical inquires, including the epistemic opacity of AI systems \citep{duran2018grounds,symons2016can,facchini2021towards,ferrario2022explainability}, fairness in machine learning \citep{barocas2021fairness,binns2018fairness},  and the nature of trust(worthiness) in human-AI interactions \citep{Floridi2019,ferrario2024justifying}, the general attitude with regard to the topic of the identity of AI systems is one of indifference. However, as recently suggested by Hatherley and Sparrow, identity and change of AI systems are an important source of societal risk \citep{hatherley2023diachronic}. For instance, in healthcare, the rapid adoption of medical AI is prompting regulators, such as the US Food and Drug Administration, to grapple with defining the boundaries of AI identity over time and its impact on predictive performance of these systems \citep{FDA2021}. Similarly, as AI systems evolve and their copies are deployed across healthcare environments with varying normative and cultural conditions, ethical challenges emerge. These include the erosion of informed consent, disruptions to patient-doctor trust, and threats to the fairness and quality of care \citep{hatherley2023diachronic}. 
The increasing pervasiveness of AI systems in society demands that philosophers, social and computer scientists, and legal scholars address the identity and evolution of these systems over time in a culturally-sensitive yet structured way. 
Our contribution holds potential to trigger these discussions and guide a metaphysics-informed development and regulation of AI.

\section{Identity, change, and identity criteria in metaphysics}
\label{section:identity_change_meta}
What is identity in metaphysics, \emph{exactly}? According to Lowe, it is ``indisputable that everything is identical to itself and with no other thing'' \citep[pag. 23]{lowe2002survey}. This is a perspective shared by a number of philosophers, although it constitutes no well-posed definition. But, then, it is puzzling that such a basic relation should even be a topic to be investigated. If we attempt to do so, intuition suggests that identity should be an equivalence relation, namely, that identity is reflexive, symmetric, and transitive. However, not much else can be said about it without falling in logical contradictions, called `puzzles,' that have characterized metaphysics since long time---for a few examples, including the celebrated `Ship of Theseus,' see Chapter 3 in \citep{gallois2016metaphysics}. These difficulties suggest that, similar to the term `existence,' 

\begin{quote}
we should accept that identity is conceptually primitive [...] and thus that it is pointless to expect an informative answer to the question `What do we \textit{mean} by \textit{identity}?' \citep[pag. 25,~emphasis in original]{lowe2015more}.    
\end{quote}

Further, continues Lowe:

\begin{quote}
[i]dentity is \textit{univocal}. There are not different \textit{kinds} of identity for different kinds of things, any more than there are different kinds of \textit{existence} -- only different \textit{criteria} of identity \citep[pag.~24, emphasis in original]{lowe2015more}.
\end{quote}

Thus, Lowe suggests that the difficulty of discussing identity in metaphysics can be overcome by introducing  rules, called \textbf{identity criteria}, to determine under which conditions two entities, for instance, two seashells found on the beach, are identical and clarify the metaphysical sense in which this identity holds. Identity criteria are rules required for the `ontological respectability' of what exists as 
``only entities which have clearly determined identity criteria are ontologically acceptable'' \citep[pag.~221]{carrara2004many}. 
\textbf{Synchronic identity criteria} formalize the identity of entities at any given time, telling us whether they are the same in a metaphysically precise sense. \textbf{Diachronic identity criteria} formalize existence over time, i.e., persistence, of entities  instead. They help addressing change over time, which occurs when a persisting entity enjoys different properties at different times  \citep{gallois2016metaphysics}. In fact, everyday perceptions suggest that plants, humans, software, and cars change around us; they undergo different types of transformation. A tree sap grows into an oak, an exercising man gains muscle mass. Software products are updated, while cars eventually begin to fail us, requiring the replacement of some  parts.
While some changes are qualitative, namely, they do not affect the persistence of the entity they refer to, others are such that this persistence stops, and a \emph{numerically} different entity emerges. In the first case, we may say that the sap and the oak are the \emph{same} plant, although enjoying different properties, while a wooden log and the ash that remains after its burning are \emph{numerically} different entities. Thus, if we were able to introduce identity criteria for plants, men, seasons, software, and cars we could argue under which transformations these entities persist, despite changing properties. In summary, using identity criteria we can \textbf{individuate, count, and investigate change over time of what exists}. It turns out that this can be done for a class of entities that we discuss in the section below.

\subsection{Entities enjoying identity criteria: Sortal terms and their kinds}\label{subsection:sortals_identity_criteria}
In metaphysics, the entities for which identity criteria  exist are called sortal terms, namely, 
``concept[s] of a distinct \textit{sort} or \textit{kind} of individuals'' \citep{lowe2015more}. The term `kind' denotes a collection of `individuals,' which may be concrete, such as pebbles, plants, cats, and humans, or abstract entities, such as mathematical objects.\footnote{Entities that do not enjoy criteria of identity are, for instance, properties and, in general, not substantival nouns, such as `red thing'  \citep{lowe2015more}.} Here, kinds and their individuals enjoy the same standing, where 
``[i]ndividuals are necessarily individuals \emph{of a kind}, and kinds are necessarily kinds \emph{of individuals}'' \citep[pag.~4, emphasis in original]{lowe2015more}.\footnote{
There are no `bare particulars,' as all individuals are necessarily instances of certain kinds and these are necessarily kinds of individuals \citep{lowe2015more}.} Thus, if an entity is an individual, such as the black cat sleeping on the couch in front of me, or my personal computer, it belongs to a kind, called \emph{cat} and \emph{personal computer} that metaphysically groups individuals of that sort. But how to determine whether an entity is an individual? The answer lies in the very definition of sortal terms, where it is stated that for any kind of individuals there is a criterion of identity for individuals of that  kind \citep{lowe2015more}. Through these identity criteria we can determine whether two individuals $x$ and $y$ in a kind $\varphi$ are the \emph{same} individual \emph{as individuals in} $\varphi$. 
Historically, Locke is the first philosopher to acknowledge explicitly the sortal relativity of criteria of identity \citep{lowe2015more}, as he recognized that `man' and `person' (both sortal terms) carry different criteria of identity \citep{locke1690essay}. More recently, it was Frege, in his \emph{Grundlagen der Arithmetik}, who explicitly mentioned the need to introduce criteria of identity of mathematical objects \citep{frege1884}. In particular, the section of the \emph{Grundlagen} comprising \S\S 62-69 states: ``To obtain the concept of Number, we must fix the sense of numerical identity.'' In \S 62, he specifies: ``[i]f we are to use the symbol $a$ to signify an object, we must have a criterion for deciding in all cases whether $b$ is the same as $a$ , even if it is not always in our power to apply this criterion.''

Let $\varphi$ denote a kind, e.g., \emph{sets}, \emph{cats}, or \emph{clocks}. Formally, a criterion of identity in $\varphi$ reads \citep{lowe2015more}:

\begin{align}
&\forall x, y \text{ in } \varphi, x \text{ is identical with } y \text{ (in symbols: } x=_{\varphi}y) \leftrightarrow \nonumber\\ 
&x \text{ and } y \text{ satisfy } R_\varphi(x,y), \label{identity_formalism}
\end{align}

where $R_\varphi(x,y)$ is an equivalence relation that depends on the kind $\varphi$.\footnote{In the literature, two types of identity criteria are usually presented: first- or second-order \citep{williamson2013identity,carrara2004many}. For a discussion of formal constraints on identity criteria, we refer to \citep[section 6.3]{carrara2014artifact}. Frege introduced a second-order criterion in \S 64 of his \emph{Grundlagen}. The criterion in \eqref{identity_formalism} is first-order instead.}
Thus, sortal kinds have different criteria of identity, as sortal terms enjoy different properties.\footnote{For instance, `man' and `gold,' says Lowe, are sortal terms as it seems that it is possible to determine whether ``if $x$ and $y$ are men or quantities of gold, they are the \textit{same} man or the \textit{same} gold'' \citep[pag.~13, emphasis in original]{lowe2015more} in the kind \emph{man} and \emph{gold}. However, the rules to identify men and quantities of gold are different. In fact, `man' is a count noun, while `gold' is a mass noun \citep{lowe2015more}.} For instance, a criterion of identity in the kind \emph{sets}, whose elements are mathematical sets reads: 

\begin{align*}
\forall A, B   \text{ sets, }   A=_{\text{sets}} B \leftrightarrow A \text{ and } B \text{ have the same elements. } 
\end{align*}

This is the axiom of extentionality in axiomatic set theory \citep{pinter2014book}. With concrete entities, one can consider

\begin{align}
\forall x, y   \text{ in } \varphi,  x=_{\varphi} y \leftrightarrow (\forall P, P(x) \leftrightarrow P(y)) \label{eq:id_crit_concrete}
\end{align}

instead. The criterion in \eqref{eq:id_crit_concrete} establishes identity through the \emph{principle of indiscernibles} (right-to-left) and the \emph{principle of the indiscernibility of identicals}---also called `Leibniz's Law'---(left-to-right), where $P$ denotes any property of individuals in $\varphi$. However, the use of \eqref{eq:id_crit_concrete} gives rise to puzzles of identity  that stimulated different perspectives on the metaphysics of concrete objects throughout time. We refer to Gallois' monograph for more details \citep{gallois2016metaphysics}. For identity criteria of humans, we suggest consulting the `Personal Identity' page of the Stanford Encyclopedia of Philosophy (SEP) instead \citep{olson2024personal}.

\vspace{0.25cm}

A couple remarks to structure the remainder of this work after these technical sections. Considerations so far suggest that committing to the idea that AI systems are sortal terms requires (1) characterizing the different kinds of AI systems, and (2) introducing the identity criteria that govern each kind (including persistence conditions). In the remainder of this paper, we aim to address this problem as follows. First, recognizing that AI systems are artifacts \citep{turner2018computational}, we start by characterizing artifact kinds \citep{carrara2014artifact,franssen2014artifact}. Next, we investigate what principles allow distinguishing different kinds of AI systems. Importantly, \textbf{we do not aim to introduce an ontology of AI systems}, namely, a formal and exhaustive classification of all different kinds of AI systems and the relations between them. Rather, our goal is to introduce a theory of identity of AI systems characterizing the different kinds of AI systems by considering the functional and socio-technical dimensions of these artifacts and  introducing the identity criteria that govern them.

\section{Artifacts and their kinds: An overview}
\label{section:ontology_artifacts}

Artifacts, such as spoons, coins, cars, smartphones, and satellites are ``objects intentionally made to serve a given purpose'' \citep[pag.~1]{baker2008shrinking}.\footnote{Note that the term `artifact' can mean different things outside the domain of  philosophy. For instance, in signal processing, an artifact refers to unintended, spurious, or undesirable alteration or distortion in a signal. In this work, we follow the philosophical perspective.}  Some are rather simple, such as spoons and coins, while others, such as cars, smartphones, and satellites, are complex systems composed of a myriad of parts that contribute to their correct functioning. Depending on the chosen account, the purpose of an artifact is realized through a function that may be characterized as technical or social, intended or accidental \citep{franssen2014artifact}. In the literature, different accounts of artifactual functions aim to capture the relationship between an artifact’s design, its use, and its purpose.  For instance, some authors propose that artifactual functions are determined by the intentions of their designers, while others argue that functions are established through the practices and goals of their users. Additionally, some scholars highlight the causal-role functions of artifacts as components within more complex systems---such as the role of a spam filter within an email management system---while others adopt etiological accounts, where a function is derived from the artifact's history of selection and use \citep{carrara2009fine}.
Regardless of their structural complexity or the functional account adopted, a common view in the study of artifacts emphasizes their dual nature: they are simultaneously material objects and functional entities \cite{baker2007metaphysics,kroes2002dual}. Their role in characterizing artifact kinds results in contrasted metaphysical perspectives that we summarize in what follows.

\subsection{The metaphysics of artifacts}\label{subsection:metaphysics_artifacts}

\subsubsection{Artifact kinds are not real kinds.} Traditionally, artifacts never enjoyed a metaphysical status comparable to the one of natural beings. According to this view, which dates back to Aristotle, chairs, ships, robots, and smartphone applications to lose weight ``may not seem to be supplied with well-defined or well-grounded \emph{persistence conditions}'' \citep[pag.~1, emphasis in original]{lowe2014real}.\footnote{Aristotle divided things into those that `exist by nature' and are `products of art,' namely, artifacts (\emph{Physica}, Book II, 192 b 28). As the art of making something, called `tékhnē,' requires intentional agency, an artifact is an object that has been intentionally made (for some purpose).} In fact, authors hold that artifacts exist only mind-dependently, that is, as products of human intentionality, lacking real essences and  `principles of activity,' such as things that exist in nature, e.g., seaweed or mammoths, do \citep{wiggins2001sameness}.\footnote{Artifacts are mind-dependent objects, although they are ``only causally mind-dependent, not constitutively mind-dependent''  \citep[pag. 20]{lowe2014real}.} In fact, seaweed and mammoths are guided by ``law-like norms of starting to exist, existing, and ceasing to exist by reference to which questions of identity and persistence can be arbitrated'' \citep[pag. 83]{wiggins2001sameness}. As a result, artifact kinds are not \emph{real} kinds \citep{wiggins2001sameness}---a famous position called \emph{antirealism} on artifact kinds \citep{wiggins1980samething}. Two consequences arise from this antirealist stance. First, artifact kinds evade precise characterization because their functions, for the antirealist, do not specify their essence as they are independent of the physical make-up of the artifact. For instance, should a folded-up chair, which cannot currently serve its function, still be considered an individual of the kind \emph{chair} \citep{wiggins2001sameness,carrara2009fine}?
In addition, it seems rather easy to individuate, count, and talk about the persistence of artifacts \citep{lowe2014real}. However, serious attempts to define the identity and persistence of objects such as chairs, excavators, and chest strap heart rate monitors lead to metaphysical puzzles due to their capacity for assembly and reassembly  \citep{gallois2016metaphysics}. Possibly, the only well-posed identity criterion that can be introduced for artifacts is `mereological essentialism' \citep{wiggins2001sameness}, which states that  artifacts are identical if and only if they share the same physical make-up. However, this criterion is too strict to be useful, as only physically immutable entities could persist.

\subsubsection{Some artifact kinds are real kinds:}
\label{subsubsection:some_artif_kinds}\emph{the function}+ \emph{approach.} Antirealism on artifact kinds has been challenged by different scholars who assert that at least \emph{some} artifact kinds can be real kinds \citep{baker2004ontology,baker2007metaphysics,elder2004real,carrara2004many,lowe2014real}. To elaborate on this point, we start by relying once again on Lowe, who contends that not all artifacts are metaphysically equal. Machines such as clocks, engines, and tractors seem to possess a \emph{unifying principle of activity} governed by `laws of action'---engineering principles that dictate how the artifact's components collectively achieve its function. These engineering principles mirror the way an organism's parts contribute to its overall functioning under a biological principle of activity \citep{lowe2014real}. Hence, Lowe commits to the idea that complex artifacts emerge as candidates for having real kinds, identity criteria, and persistence conditions. Additionally, several scholars argue that, regardless of the functional account adopted, artifact functions should be considered an integral \emph{part} of their essence. This view provides a foundation for discussing artifact kinds and their identity criteria. As Kornblith puts it: ``At least for the most part, it seems that what makes two artifacts members of the same kind is that they perform \emph{the same function}'' \citep[p.~112, emphasis ours]{kornblith1980referring}. Thus, endorsing the realism of (some) artifact kinds requires the conjunction of functions with other features of artifacts  \citep{carrara2014artifact}.\footnote{ A first example of conjunction comes from Wiggins \citep{wiggins2001sameness}. He states that, given an artifact $a$ at time $t$ and an artifact $b$ at time $t'$, if $a$ and $b$ are in $\varphi$, then $a=_\varphi b \leftrightarrow$ there is a continuous material path between $a$ and $b$ and $a$ and $b$ perform the same function \citep{wiggins2001sameness}. More details in \citep{carrara2014artifact}.} 

An interesting type of conjunction is Carrara and Vermaas' `\emph{function}+' \citep{carrara2009fine}. It characterizes artifact kinds via artifactual functions: two artifacts belong to the same kind if they have the same function. Further, it introduces identity criteria for artifacts as the conjunction of (1) \textbf{their function}, (2) an `\textbf{operational principle},' and (3) their `\textbf{normal configuration}.' The operational principle of an artifact are \emph{rules} that specify ``how characteristic parts [of the artifact]...fulfill their special function in combining to an overall operation which achieves the purpose [of the artifact]'' \citep[pag.~328]{polanyi2012personal}. In essence, the operational principle reflects Lowe's idea of a  unifying principle of activity for complex artifacts, where the artifact's parts---e.g., devices, interfaces, hardware, or software---work together under engineering norms to perform the artifact's function \citep{lowe2014real}. The design of each part is specified by a  `normal configuration,' instead. This concept refers to the 
``general shape and arrangement [of the artifact parts] that are commonly agreed to \emph{best} embody the operational principle'' \citep[pag.~209-210, emphasis ours]{vincenti1990what}. 
To embody the operational principle at best, these configurations are typically derived from design principles, industry standards, and best practices in different  engineering disciplines. They serve as reference models for how a system should be assembled, operated, or maintained to ensure it performs its intended function---an important point that will return in Section \ref{subsection:TW_AI}. Patents are typically used as the standards that formulate the operational principle of complex artifacts \citep[pag.~345]{vincenti1990what}.
However, they do not specify normal configurations, as ``the inventor of a machine will always try to obtain a patent in the wildest possible terms'' \citep[pag.~345]{vincenti1990what} with no reference to the particulars of any constructed machine.
 
In summary, Carrara and Vermaas' \emph{function}+ promotes the realism of  artifact kinds by combining the functions of these objects with their internal structures and engineering principles. Endorsing \emph{function}+, interesting consequences follow. First, as noted above, two artifacts having different functions are not in the same kind. Then, artifacts made to fly, let us say airplanes and rockets belong to the same artifact kind---that we may call \emph{flying artifacts}. However, the identity criterion defined by \emph{function}+---referred to as `technological identity' by Carrara and Vermaas \citep{carrara2009fine}---asserts that an airplane and a rocket cannot be considered the same artifact within the category of \emph{flying artifacts}. This distinction arises because these artifacts adhere to fundamentally different operational principles and configurations. (From an ontological perspective, one might argue that the category of \emph{flying artifacts} can be further refined into subcategories such as \emph{airplanes}, \emph{rockets}, and others.) 
Finally, the technological identity criterion derived from \emph{function}+ is ``more liberal'' \citep[p.~134]{carrara2009fine} than mereological essentialism. Specifically, under \emph{function}+, \textbf{two artifacts belonging to the same kind may be identical despite their physical structures being different}. Thus, despite potential puzzlement by some, these artifacts have the right to be regarded as being identical within their kind. Similar considerations hold for the persistence of artifacts. We argue that the \emph{function}+ framework compellingly emphasizes the necessity of a functional approach to the metaphysics of artifacts \citep{carrara2004many,lowe2014real}. By treating complex artifacts as products of engineering, it establishes a connection between their function and physical configurations across their artifact histories. We will endorse this approach for AI systems opening the door to metaphysical considerations that we explore in what follows.

\section{Metaphysics of AI systems: Their kinds and identity criteria}
\label{section:identity_change_AI}

\subsection{What are AI systems?}
\label{subsection:what_are_AI}

\subsubsection{Different shapes and forms of AI systems.} AI systems are computational artifacts, namely, artifacts of computer science \citep{turner2018computational}. These systems typically consist of both software---such as machine learning models\footnote{In this work, we consider (trained) machine learning models as \emph{proper parts} of AI systems. These models are artifacts, but not AI systems.}---and hardware, which includes different devices, sensors, interfaces, and data processing units. They come in a notable variety of forms, with diverse structures and capabilities. Large language model (LLM)-based conversational agents, such as ChatGPT, are AI systems that interact with users through apps or web-based applications. AI-powered decision-support systems include medical devices that monitor patients’ physiological signals in real-time, predicting risks such as sepsis or heart failure. Recommendation systems used by e-commerce platforms tailor content to users using machine learning models. Fraud detection systems, virtual assistants, and social robots are additional examples of modern AI systems. Recent systematic reviews, such as \citep{benk2024twenty,Vereschak2021,Knickrehm2023}, have begun to highlight evidence-based branches of a taxonomy for AI systems. However, they do not attempt to formalize one.

\subsubsection{Machine learning retraining as redesign of AI systems.}\label{subsubsection:ML_retrainings} Despite this variability of shapes and forms, something can be said that is valid for all AI systems: these systems are dynamic to some extent, with components that can change over time due to software updates or hardware replacements, among others. A crucial form of update is the \textbf{retraining of the machine learning models} that AI systems use to compute their outputs—such as scores, labels, images, or text. Retraining is essential to maintain the accuracy of these models in the face of changing environments and data patterns.\footnote{For instance, customer preferences may evolve, healthcare providers may introduce new treatments, and new types of input data become available. Designers must accommodate these changes by updating the machine learning models to ensure that the system continues to perform effectively.} Retraining can involve different redesign procedures that differ significantly across different types of AI systems, in terms of both the frequency and the resources required. They may involve optimizing hyperparameters within the same model family using new data, switching to different model families, or redefining the statistical learning problem by, for instance, moving from binary classification to multi-class predictions. As a result, retraining does not only affect the accuracy of the AI system, but also other properties that are important for the correct functioning of the system. In fact, it can help enforcing fairness constraints, improving explainability by refining decision boundaries, reducing unexpected behaviors that are hard to explain, or using more understandable features. Further, retraining may necessitate changes to the physical and logical configuration of the system, such as adding or removing interfaces and devices or modifying user functionalities based on the updated characteristics of the redesigned model. In this work, we adopt a broad interpretation of  `retraining,' defining it as \textbf{a redesign procedure for an AI system centered around updating its machine learning models}, which influences both the system’s functionality and, potentially, its physical make-up.\footnote{Redesign procedures that do not fall under the retraining umbrella comprise general user interface redesigns, energy efficiency improvements, and other form of hardware optimization that do not affect trained machine learning models.} 

\subsection{Are AI system kinds real?}\label{subsection:AI_kinds}
We examine whether AI system kinds are real kinds by adopting the \emph{function}+ approach discussed in Section \ref{subsection:metaphysics_artifacts}. To do so, we characterize AI system kinds by analyzing their functions, operational principles, and normal configurations. A successful outcome of this analysis would allow us introducing identity criteria of AI systems using Carrara and Vermaas' technology identity criterion straightforwardly \citep{carrara2009fine}. As will become evident, we will not succeed in our quest.

\subsubsection{AI system functions and their specifications}
The most widely used definitions of `artificial intelligence' refer to the design of artifacts that \emph{attempt simulating} aspects of human intelligence \citep{RussellNorvig2020,Turing1950}. Thus, at the highest level of abstraction, the function of AI systems is to perform activities that attempt simulating cognitive and epistemic capabilities humans use to process and manage information, enabling tasks such as monitoring other systems, solving problems, and supporting decision-making. This function can be categorized into three  types: \emph{prediction}—the computation of numerical scores or labels for classifying data points or selecting optimal actions based on computational criteria, \emph{pattern recognition}—labeling data by identifying similarities between data points, and \emph{generation}—producing new data in a specific format, such as textual strings, images, or audio samples, a capability particularly prominent in multimodal LLM-based applications.
This level of functional description, however, is too coarse to provide a meaningful framework for understanding the identity of AI systems. Such broad categorizations could group vastly different systems---such as medical AI predicting risk scores, matchmaking algorithms assessing compatibility, facial recognition tools, and military targeting software---under the same kind, namely, \emph{predictors}. Similarly, systems performing pattern recognition or generation would be grouped under equally broad categories. This is analogous to categorizing both chairs and tables as `four-legged objects supporting other concrete entities.' While technically accurate, such a level lacks practical relevance for studying identity criteria or addressing real-world applications.\footnote{Of course, AI systems can perform combinations of these high-level functions, as seen in LLM-based conversational agents. This possibility does not undermine our argument, as it neither depends on nor seeks to establish an ontology of AI systems.}
A more refined approach is to express the function of an AI system using formulations such as `[to predict/recognize/generate] \texttt{X} using \texttt{Y},' where \texttt{X} represents the object of the activity and 
\texttt{Y} the resources used to perform it.\footnote{ For instance, we may state that the AI system $S$ performs the function of 
\begin{enumerate}
\item  predicting the output $X$ (e.g., a numerical score) using data $Y$ (e.g., sociodemographic and financial information),
\item predicting the output $X$ using data $Y$, and model $Z$ (e.g., a feedforward neural network with $N$ neurons and $M$ layers),
\end{enumerate}
and so on. This progressive refinement parallels the classification of physical objects in ontology.} 
We refer to this level of functional specification as the `techno-function' of the AI system, capturing the artifact's technical capabilities as intended by its designers. The techno-function emphasizes what the system is designed to achieve and how it accomplishes its objectives through specific inputs, models, and routines. As with physical artifacts, the level of detail in techno-function specifications is not fixed and can vary depending on theoretical or practical considerations.
In general, refining the specification of a techno-function narrows the corresponding kind of artifacts capable of performing it. This process leads to a sequence of nested AI system kinds, denoted as $\varphi_1 \supset \varphi_2 \supset \dots \supset \varphi_n \supset \dots$, where each subsequent kind is characterized by an increasingly detailed techno-functional description. Since identity criteria are relative to the chosen kind, we can select any level of specification within this hierarchy, examine the corresponding kind of AI systems, and proceed to analyze their operational principles and normal configurations.

\subsubsection{Operational principles and normal configurations of AI systems.}\label{subsubsection:op_principle_normal_config_AI}
By definition, the operational principle of an AI system consists of the rules that specify how its components fulfill their operations to achieve the system's purpose, as defined by its techno-function. Crucially, there is no unique operational principle that universally applies to all AI systems performing the same techno-function. This lack of uniqueness arises from the diverse applications of AI systems, particularly in high-risk domains, and their reliance on machine learning. In fact, the conditions that determine the \emph{correct functioning} for these systems are strongly influenced by the norms and practices of the social systems in which they are designed and deployed. Normative expectations affecting AI systems are notoriously complex, including, for instance, considerations of performance, transparency, and fairness affecting the design of their machine learning models, as well as inclusion (e.g., for users with disabilities) or control over their users. This list is not exhaustive. \textbf{Altogether, these considerations contribute shaping their logical and physical configurations throughout their artifact histories}. After all, as noted by Lowe: 
``[t]he existence and identity conditions of the members of artefact kinds plausibly carry reference to human or other intelligent \emph{artificers} and \emph{utilizers} of them'' \citep[pag. 20, emphasis in original]{lowe2014real}. Thus, unlike much simpler  artifacts, such as spoons, or nails, we contend that there is no universal operational principle for medical AI systems, spam filters, AI-powered surveillance tools, or LLM-based conversational systems. Rather, operational principles are relative to the social system where these systems are designed and deployed and can be modified throughout the artifact history.

Further, while common artifacts have well-defined normal configurations, or admit few variations of them, such as electric kettles, AI systems typically fail those. The operational principle of an AI system is implemented in different ways in a social system. Components of AI systems can be specified through engineering standards, such as patents or industry guidelines. For instance, the US Patent US10191537B2 describes customized haptic feedback for wearable devices, while the ISO/IEC 24775 series provides standards for data storage management.
However, many AI systems, such as spam filters, LLM-based conversational agents, sepsis predictors, or deepfake generators, lack patents or industry standards, even though some of their individual components may adhere to them. Additionally, the use of services, such as Google’s AutoML,\footnote{\url{https://cloud.google.com/automl}} further promotes building custom AI systems without adhering to specific engineering standards, but, rather, pragmatic design aspects.

\vspace{0.25cm}

In summary, developing a theory of identity for AI systems based on the \emph{function}+ framework presents some challenges. While the techno-function seems to be the appropriate functional level for characterizing AI system kinds, these artifacts present relativized operational principles and, in general, lack standardized normal configurations. In the following section, we address these challenges using a key concept in the ethics of AI: \textbf{trustworthiness}. It will help us bridging the gap between the techno-functional characterization of AI systems and the absence of universal artifact principles and their standards.

\section{Our approach: A trustworthiness-based  metaphysics of AI systems}
\label{section:our_approach}

\subsection{The trustworthiness of AI systems}\label{subsection:TW_AI}
Quite literally, the trustworthiness of AI systems is their capability to be `worthy of our trust.' Although trust is a notoriously challenging construct to define---one that has engaged philosophers in conceptual debates for a long time (for an overview, see the excellent SEP entry `Trust' \citep{sep-trust})---most scholars agree that it is possible to conceptualize the trustworthiness of AI systems. This is typically achieved by adopting a \emph{principled} approach. In such an approach, different actors within a social system identify principles that technology should uphold to mitigate various risks and steer the development of AI systems in alignment with societal norms and practices.\footnote{Here, `social system' can refer to different types of organizations expressing principled considerations on what makes AI worthy of our trust, including research groups, private companies, non-profit organizations, and supranational entities \cite{EuropeanCommission2018,GoogleAIPrinciples,AlgorithmWatchAIEthicsInventory, jacovi2021formalizing,Loi2019}.}
A well-known example comes from the European Commission’s High-Level Expert Group on Artificial Intelligence, which, in 2018, proposed that trustworthy AI should be lawful, ethical, and robust. This framework emphasizes principles such as human agency, technical robustness, safety, and transparency \citep{EuropeanCommission2018}. These principles are then translated by designers into system capabilities, including accuracy, fairness, safety, transparency, and technical reliability \citep{Floridi2019,Li2023}.
Collectively, these capabilities constitute the trustworthiness of AI, which, in turn, is referred to as a second-order---or `composite'---capability of AI that is expression of the normative and practical demands of a social system \citep{Kaur2022,Mattioli2024,ferrario2024justifying,Floridi2020}. Further, the trustworthiness of an AI system is context-dependent. In fact, not all AI systems---even in the same social system---require the same degree of accuracy, fairness, or explainability to function as intended or, at least, to the same degree. For instance, while providing detailed yet comprehensible explanations of a medical AI's predictions is considered critical in intensive care contexts, this requirement is less pertinent for machine learning-based spam filters. In personalized marketing, the accuracy of recommendations may be less critical than it is in AI-supported airport traffic monitoring routines. 

The need for contextualizing the trustworthiness of AI systems has prompted Jacovi et al. to define trustworthiness as the capability of an AI system to fulfill explicit, predefined commitments, referred to as \textbf{contracts} \citep{jacovi2021formalizing}---an approach that stems from Hawley and Talliant works on `contractual trust' \citep{Hawley2014,tallant2017commitment}. These contracts are the capabilities of the AI system that constitute its trustworthiness, such as predictive performance, fairness and technical reliability, and are typically specified by the system's designers, evaluated, and enforced during retraining routines. 
For instance, in their work, Jacovi et al. explicitly mention ``model correctness,'' i.e., accuracy, and ``transparency'' as genuine examples of AI contracts \citep{jacovi2021formalizing}. Typically, an AI system is required to maintain different contracts, i.e., one per system capability (as in the examples above).
In summary, trustworthiness can be understood as the  capability of an AI system to uphold predefined commitments that ensure its \emph{correct functioning} \citep{Kaur2022,Mattioli2024,ferrario2022explainability}. In the terminology of the \emph{function}+ approach, the \textbf{trustworthiness of an AI system encodes its operational principle}. It gives us reasons to talk about a unifying principle of activity for these artifacts that is expressed via the rules dictated by the properties that constitute it \citep{lowe2014real}. 
We define the collection of contracts that formulate the trustworthiness requirements of an AI system its `\textbf{trustworthiness profile}.' Just as the operational principle of traditional artifacts is  formalized through patents (see Section \ref{subsubsection:op_principle_normal_config_AI}), the trustworthiness of AI systems is articulated through contracts. These contracts---along with accompanying documentation specifying evaluation methods---outline the configurations of the system's components necessary to fulfill its ascribed techno-function. At a first glance, all configurations, both logical and physical, that satisfy the contracts the system is required to uphold could be considered `normal.' In practice, there exist multiple implementations that can fulfill the same contract. For instance, `maintain accuracy above 90\%' can be achieved through different machine learning models. 
Similarly, a fairness-related contract, such as `limit the false positive rate across demographic groups to a maximum of 7.5\%,' might be satisfied through preprocessing training data or applying post-hoc adjustments to model outputs.\footnote{By increasing the level of specification for a contract, it is theoretically possible to narrow the range of normal configurations that satisfy it. However, we argue that no feasible level of specification can uniquely determine a single normal configuration for every contract. Achieving this would require imposing detailed constraints on every stage of the machine learning pipeline, including data distributions and their statistical properties under environmental shifts, preprocessing steps, feature engineering, and model selection. Such exhaustive requirements would result in contracts that are impractical to manage and overly rigid, hindering the adaptability of AI systems to real-world scenarios over time.} 
There is no single \emph{best} way to satisfy contracts, as required by the definition of `normal configuration'---see Section \ref{subsubsection:some_artif_kinds}. Instead, there exist (equivalence) classes of normal configurations that realize the trustworthiness profile of an AI system with equal efficiency.

%

\begin{figure}[h!]
    \centering
    \includegraphics[width=0.5\textwidth]{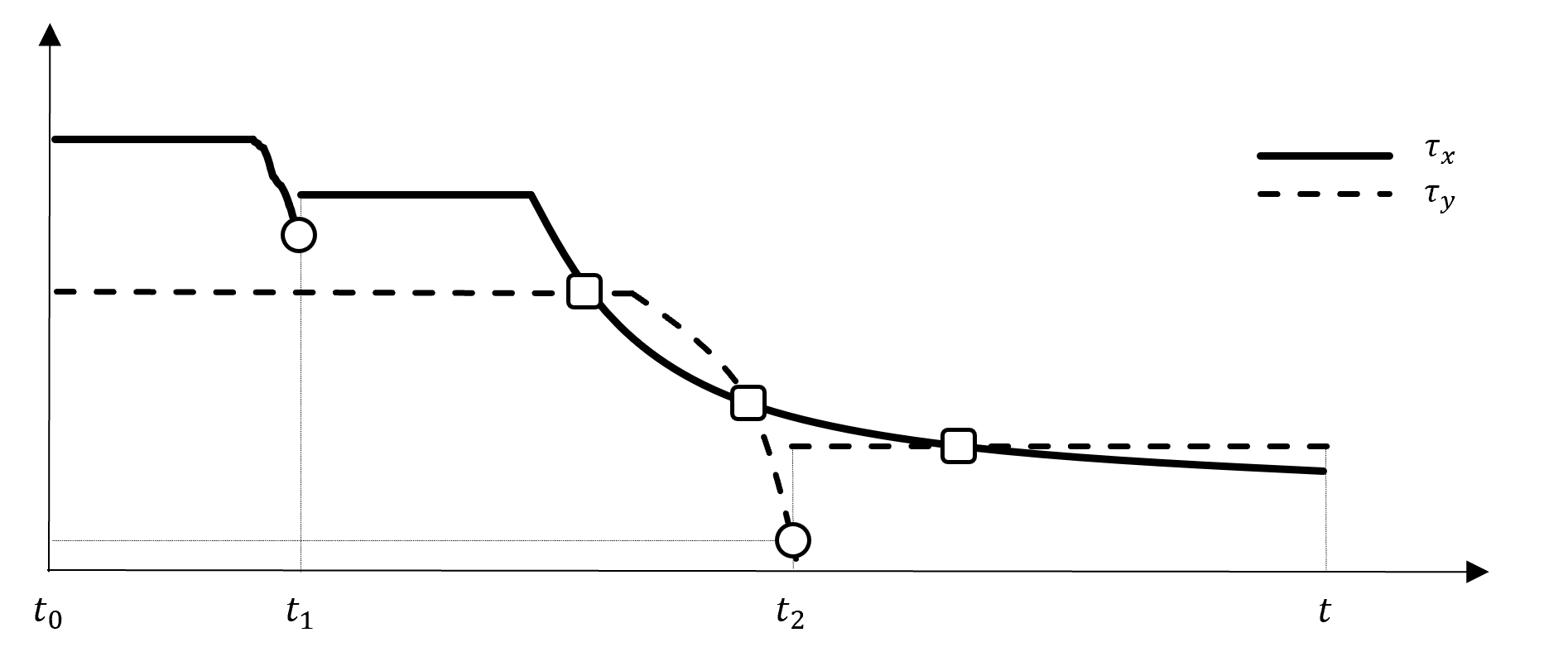}  \caption{Levels of trustworthiness over time of two AI systems $x$ and $y$ in a kind $\varphi$. They require retraining at $t_1$ and $t_2$ due to rapid decreases in their trustworthiness levels. With reference to synchronic identity---see \eqref{eq:synchro}---$x$ and $y$ can be deemed identical at the times indicated by squares. With reference to diachronic identity---see \eqref{eq:diachro}---$y$ can persist through the interval $[t_2,t]$.}
    \label{fig:trustworthiness_levels}
\end{figure}

Following the discussions in \citep{ferrario2024justifying}, we formalize this efficiency  by means of a non-negative function $\tau_x$, where $\tau_x(t)$  denotes the trustworthiness level of the AI system $x$ at time \(t\). By definition, \(\tau_x(t)\) is a function of the levels of the individual capabilities---e.g., accuracy---that \(x\) is required to uphold according to its contractual obligations over time. The hypothesis  is that each such capability has levels---also depending on time---that can be formalized by means of its measurability \citep{ferrario2024justifying}. (Different capabilities, e.g., accuracy and fairness, are measured with different instruments by system  designers---see \citep{ferrario2024justifying}.) For instance, in a toy AI system $x$, trustworthiness is equated to accuracy with the contract  `maintain accuracy above 90\%.' Then, 
$ 
\tau_x(t) =
\begin{cases} 
1 & \text{accuracy}(t) \geq 0.9, \\
0.5 &  0.7 \leq \text{accuracy}(t) < 0.9, \\
0 & \text{otherwise.}
\end{cases}
$
is an example of trustworthiness function. Note that these levels are constant over \emph{intervals} of predictive performance.
Let \(t_0^x\) denote the time at which \(x\) is successfully deployed in its operational environment. In practice, some of the capabilities of the system $x$ may start to degrade and even invalidate their corresponding contract at some time $t\geq t_0^x$. Eventually, some properties may decrease to the level that, overall, the trustworthiness levels \(\tau_x(t)\) degrade as well, with the rate and pattern of degradation depending on the specific context in which \(x\) operates and the definition of $\tau_x$. To mitigate this decline, designers periodically retrain the machine learning models in the system, as discussed in Section~\ref{subsubsection:ML_retrainings}. These retraining events partition the AI system's life cycle  into consecutive time intervals, with the endpoints of these intervals coinciding with retraining routines. At these points, the trustworthiness function \(\tau_x\) exhibits \emph{discontinuities}, as illustrated in Figure~\ref{fig:trustworthiness_levels}.

Before introducing our identity criteria, we make a final observation about contracts and the function $\tau_x$. While Jacovi et al. adopt high-level formulations for contracts, these could also be specified in greater detail. For instance, a contract might specify `maintaining accuracy above 90\%' (accuracy) or ensuring the system `maintains a 99.9\% uptime over any 30-day period, with no single downtime exceeding 10 minutes' (technical reliability). However, there is no universal standard for how contracts should be formulated. Recommendations focus on what system capabilities should be included in contracts \citep{EuropeanCommission2018}, while evaluation methodologies and standardized documentation remain active areas of research---see \citep[\S 3.2]{jacovi2021formalizing}. Similar considerations hold for $\tau_x$ whose formulation may follow theoretical and practical considerations \citep{ferrario2024justifying}. Then, in the forthcoming discussions, we assume that whenever the identity or persistence of artifacts $x$ and $y$ is considered, (1) a common level of formulation is established for the contracts that constitute their trustworthiness profiles, and (2) the trustworthiness functions $\tau_x$ and $\tau_y$ are expressed using the same formulation. These regularity assumptions are reasonable, for instance, in social systems producing AI systems where standardized procedures are already in place or could be achieved through enforceable policies.

\subsection{Trustworthiness-based identity criteria of AI systems} \label{subsection:TW_AI_criteria}
We are now ready to introduce the identity criteria for AI system kinds at the core of our approach. While their general form are provided in the Appendix, we present here their two perspectives: \textbf{synchronic identity} and \textbf{diachronic identity}. Some notation: let $x$ and $y$ be AI systems of kind $\varphi$.  Denote the state of $x$ at any time $t'$ by $x(t')$, and similarly for $y$.

\begin{definition}[\textbf{Identity Criteria for AI System Kinds}] Let $x$ and $y$ be AI systems of kind $\varphi$ such that $t^x_0 = t^y_0 := t_0$.

\textbf{(1)} The criterion for \textbf{synchronic identity} in $\varphi$ is:

\begin{align} 
x(t) =_{\varphi} y(t) \leftrightarrow~~  & x(t) \text{ and } y(t) \text{ are individuals of kind } \varphi, \nonumber \\ 
& x(t) \text{ and } y(t) \text{ have equal} \nonumber \\ & \text{trustworthiness profiles}, \text{ and } \nonumber \\ 
& \tau_x(t) = \tau_y(t),~~ t\geq t_0. \label{eq:synchro} 
\end{align}

\textbf{(2)} The criterion for \textbf{diachronic identity} in $\varphi$ is:

\begin{align} 
x(t_1) =_{\varphi} x(t_2) \leftrightarrow~~ & x(t_1) \text{ and } x(t_2) \text{ are individuals of kind } \varphi, \nonumber \\
& x(t_1) \text{ and } x(t_2) \text{ have equal} \nonumber \\ & \text{trustworthiness profiles}, \text{ and } \nonumber \\ 
& \tau_x(t_1) = \tau_x(t_2),~~ t_1,t_2\geq t_0. \label{eq:diachro} 
\end{align} 
\end{definition}

In the Appendix, we provide an analysis of these criteria, including their equivalence conditions and underlying rationale. Here, we briefly state the key points: two AI systems belonging to the same kind are identical at any given time if and only if they execute the same techno-function, adhere to identical operational principle specifications, and have equal trustworthiness levels at that time (\textbf{synchronic identity}). An AI system persists between times $t_1$ and $t_2$ if and only if it belongs to its kind at $t_1$ and $t_2$ with the same  trustworthiness profile and levels at those times (\textbf{diachronic identity}). In Wiggins' terminology \citep{wiggins2001sameness}, conditions \eqref{eq:synchro} and \eqref{eq:diachro} rely on continuous in time `functional and operational' paths. For instance, \eqref{eq:diachro} requires such a path connecting   $x$ at $t_1$ and $x$ at $t_2$ with `equal' endpoint conditions. Beyond the metaphysics of artifacts, the concept of `path' is also utilized in the philosophy of mind, where (psychological) paths are employed by some to define personal identity---see \citep{olson2024personal}.  As a result, $=_{\varphi}$ is a trustworthiness-based technology identity criterion in the sense of Carrara and Vermaas \citep{carrara2009fine}.

\subsubsection{Identity of AI systems: some examples.} We move to a few examples of identity scenarios for AI systems.
\paragraph{Identity of toy systems.} Let  $x$ and $y$ denote two toy AI systems designed by Bob to predict numerical scores on some tabular data. Since their deployment, the systems share the same techno-function and trustworthiness profiles, which coincides with accuracy---the only system property Bob cares about. While $x$ operates as a locally run Jupyter Notebook used by Bob as a sandbox, $y$ is deployed in Google Colab using Bob's account. Despite differences in execution environments (e.g., on-premises vs. cloud), these systems are identical at a given time if their accuracy is equal. However, the accuracy of the system $x$ begins decreasing monotonically after a certain time, let us say $t$, ceteris paribus. Thus $x(t_1) =_{\varphi} x(t_2)$ if false for all $t_1$ and $t_2$ such that $t_1<t<t_2$.

\paragraph{Identity in stable operational environments.} Two AI-powered machines begin operations on the same day in a warehouse, tasked with sorting parcels onto predefined tracks. Each machine uses a different deep learning model for classifying parcels and controlling the track switches. Suppose the two AI systems perform the same techno-function and maintain equal trustworthiness profiles since deployment. Further, they operate under constant trustworthiness levels over time due to stable environmental conditions. In this scenario, the machines are considered identical at any time $t$ if and only if their trustworthiness levels are equal at $t$. This means that, despite potentially differing levels of accuracy, reliability, or safety, due to different logical and physical make-ups, overall, they satisfy their operational principle \emph{equally} and they are deemed identical as sorting robots. Given the notable differences in their design and maintenance costs, management finds this argument difficult to accept.

\paragraph{Faulty copies are not identical to well-functioning ones.} Inspired by Hatherley and Sparrow's work \citep{hatherley2023diachronic}, consider 
$x$ and $y$ as two copies of an AI-powered diagnostic tool sold by the company `Bone Breaks AI' to two hospitals for detecting small bone fractures in X-ray images. Upon deployment in the same day, both tools show the same trustworthiness profiles and levels.  However, over time, the hospitals adopt different machine learning operations procedures to maintain and improve the tools. In one hospital, 
$x$ undergoes frequent retraining with carefully curated and diverse datasets, following a rigorous evaluation routine. Meanwhile, in the other hospital, 
$y$ is retrained less frequently, using smaller or less representative datasets, and without robust validation processes. As a result, $x$ consistently outperforms 
$y$ in terms of trustworthiness, with superior accuracy, reliability, and robustness after some time from their deployment. Despite what an orthopedic surgeon using $y$ at the second hospital might tell a patient who received a different prediction from $x$ at the first hospital, the two systems are not identical. Further, the type of machine learning operations at the second hospital lead to the interruption of $y$'s persistence. 

\paragraph{One or two?}
A tech-company develops an LLM-based conversational assistant to provide users with `knowledge pills,' bite-sized insights for everyday use. The assistant  answers general knowledge queries, such as recalling historical facts, operating with a consistent trustworthiness profile, while maintaining high answer relevancy and contextual relevancy in its responses. The assistant is deployed in a country, serving its users efficiently. One day, the company's CEO surprises everyone at a board meeting announcing that the system must be deployed in another country in a more politically sensitive region, where new regulatory requirements mandate that AI systems adhere to stricter constraints, including compliance with local laws on acceptable language, censorship of politically and historically sensitive topics, and constraints to address socio-cultural biases. To adhere to these requirements, a team of engineers deploys a second instance of the AI system in the new country, modifying the system by
fine-tuning its LLM with region-specific data, implementing content moderation algorithms, and adding what they referred to as `fairness constraints' in a newly introduced `user safety' contract. 
In a heated meeting on resource allocation for the engineering team, the CEO and the legal department insist that the \emph{second} system is `just another version' of the \emph{original} system, emphasizing that both run on the same infrastructure. Engineers argue that the technical changes are such that they face two \emph{numerically} different systems instead. Metaphysical considerations may support their claim.


\paragraph{Sometimes a new AI product is new.}
A fitness company develops an AI-powered app  designed to provide personalized workout plans and real-time feedback based on users' fitness data. Since its deployment, the app has maintained a stable trustworthiness profile with constant levels of accuracy, safety, and technical robustness. However, months after release, data engineers discovered a bug in its machine learning pipeline that caused a systematic bias---users with certain characteristics were consistently receiving less effective workout suggestions. As a result, the app's trustworthiness levels began to decline. To address the issue, engineers opted for a major overhaul: replacing the original model with a more advanced one, using newly engineered features for better personalization, but keeping the system's trustworthiness profile. A few tests demonstrated that the app's trustworthiness had returned to its original level. Encouraged by the results, the marketing team proposed relaunching the app as a brand-new version and asking its users a higher subscription fee. This seems to be justified at least metaphysically, as the system did not persist.

\paragraph{The AI of Theseus? On the intermittent existence of AI systems}
Endorsing the \emph{function}+ approach, our theory of identity for AI systems avoids puzzles of identity, such as the Ship of Theseus, which in their classical formulations focus solely on the physical components of an artifact while neglecting its functional continuity \citep{hobbes1655dec,lowe1983identity}. However, as stated in \eqref{eq:diachro}, if the trustworthiness levels of an AI system drop after deployment, the system eventually ceases to persist, and a numerically different AI system emerges. Does this imply that AI systems flicker in and out of existence? We argue that this is not the case. First, the function $\tau_x$ can be meaningfully defined within a social system in a \textbf{robust} manner, ensuring it remains stable under (small) variations in the levels of the capacities that constitute the trustworthiness of system $x$. Second, a change in trustworthiness levels can be seen as analogous to disassembling an artifact to the point where it can no longer perform its function. For instance, while the collection of gears  on a table \emph{constitutes} the watch that was once on my wrist, it is not identical to it. Constitution and identity are fundamentally distinct relations \citep{gallois2016metaphysics}. Only when the parts are reassembled in a way that supports the watch's function does a watch emerge. However, this is a different watch constituted by the parts that were once mine and the duration of the interval during which the watch was disassembled has no bearing on its identity.  Similarly, retraining an AI system to restore its trustworthiness levels ensures the system's correct functioning and allows it to persist once again as a system that may \emph{only} share the physical make-up of the \emph{original}.  
\section{Conclusions}\label{section:conclusions}
We introduced a theory of identity for AI systems grounded in the \emph{function+} approach, providing rules to individuate and count AI systems as well as  to discuss their persistence. Central to our approach is the concept of trustworthiness, which relates functional expectations with the logical and physical make-ups of these artifacts. Our theory emphasizes that the metaphysical identity of AI is intertwined with  the norms and practices of the social systems in which  these artifacts are designed and deployed. Thus, the implications of retraining machine learning models, imposing fairness constraints, and measuring trustworthiness on the identity and persistence of AI systems hinges on socio-cultural perspectives, ethical and legal guidelines, as well as practical considerations specific to each of these social systems. Our theory underscores the need for standardization in these practices, including the adoption of reproducible methodologies to document trustworthiness contracts and measure AI capabilities consistently over time beyond, for instance, generic transparency requirements, such as those in the EU AI Act--see preamble 71 therein. By doing so, we can ground epistemological, ethical, and legal discussions about the present and future of AI in clearer metaphysical foundations, avoiding the pitfalls of semantic ambiguity. Armed with our theory of identity of AI systems, we can answer the \emph{epistemic} question: `If $x$ and $y$ are in the kind $\varphi$, how can we \emph{know} that $x$ is the identical to $y$?' \citep{carrara2014artifact}. Further, the ethical and legal aspects of questions such as `Is \emph{the} system we use with this patient the \emph{same} as one month ago?,' `Is the \emph{other} system better than \emph{this one}?,' or `Is \emph{the} system now less opaque, more trustworthy, or hallucinating?' can then be addressed with the clarity they deserve.

\begin{acks}
I am grateful to Michele Loi for helpful  feedback on an earlier draft of this manuscript. I also thank the anonymous reviewers for their constructive comments, which helped strengthen the final version. This work was supported by the Swiss National Science Foundation (SNSF), grant no. 229061.
\end{acks}

\section*{Appendix}
We introduce our identity criteria for the AI system kinds that follows the \emph{function}+ approach. We comment on its conditions and show how to retrieve synchronic and diachronic identity---see \eqref{eq:synchro} and \eqref{eq:diachro} in Section 5.2---from it.

\begin{definition}[\textbf{Identity Criteria of AI Systems}]
Let $x$ and $y$ be AI systems of kind $\varphi$ such that their deployment times satisfy $t^x_0 = t^y_0 := t_0$. Consider $x$ at time $t_1$ and $y$ at time $t_2$, denoted by $x(t_1)$ and $y(t_2)$, respectively. The identity criterion in $\varphi$ is defined as:

\begin{align*}
x(t_1) =_{\varphi} y(t_2) \leftrightarrow  R_\varphi(x(t_1), y(t_2)),
\end{align*}

where $R_\varphi(x(t_1), y(t_2))$ holds if and only if the following conditions are satisfied:

\begin{align}
&  x(t_1) \text{ and } y(t_2) \text{ are individuals of kind } \varphi,  \label{eq:kind}\\
&  x(t_1) \text{ and } y(t_2) \text{ have equal trustworthiness profiles, and}  \label{eq:tw} \\
&  \tau_x(t_1) = \tau_y(t_2),~~t_1,t_2 \geq t_0. \label{eq:tw_level}
\end{align}

\end{definition}

\paragraph{A few remarks on $=_{\varphi}$.} Condition \eqref{eq:kind} stipulates that the artifacts whose identity is to be checked are individuals of the kind $\varphi$.\ Condition \eqref{eq:tw} enforces that the operational principle of both systems is the same, despite possible system retraining and different physical make-ups. As mentioned in Section \ref{subsection:TW_AI}, we suppose that the trustworthiness profiles of $x$ and $y$ are formulated with the same level of specification. (Note that this relation is indeed an equivalence relation.) Thus, our identity criteria can be applied to AI systems developed within social systems that share comparable standards for trustworthiness requirements. Finally, condition \eqref{eq:tw_level} specifies that the operational performance of the systems must be equal at the times specified in the relation for identity in $\varphi$ to hold. Altogether, $=_\varphi$ stipulates that artifacts showing the same  techno-function and operational principle, including its quantification, are deemed identical in $\varphi$. In summary, $=_{\varphi}$ constitutes a trustworthiness-based technological identity criterion in the sense of Carrara and Vermaas \citep{carrara2009fine}. This criterion is more liberal than mereological essentialism, as it allows identical artifacts to differ in their physical make-up. 

\paragraph{Is $=_\varphi$ an equivalence relation?} It is not difficult to check that $=_{\varphi}$ is an equivalence relation. Reflexivity and symmetry from the definition of $=_\varphi$.  Transitivity is enforced by having equal trustworthiness profiles at the times considered in the identity criterion. 
However, transitivity is easily violated when attempting to compare trustworthiness levels directly using, for instance, distance-based conditions  between $\tau_x(t)$ and $\tau_y(t)$ over intervals such as $[t_1, t_2]$ or $[t_0, \max(t_1, t_2)]$, due to the triangular inequality of distance functions. Condition \eqref{eq:tw_level} ensures transitivity while relating operational performance of the artifact in an easy way. More mathematically complex conditions on the trustworthiness levels, while potentially preserving transitivity, tend to lose their connection to the \emph{function}+ framework and become less justifiable. 

\vspace{0.25cm}

Finally, when $t_1 = t_2 = t$, $=_{\varphi}$ provides a synchronic identity criterion for AI systems---see \eqref{eq:synchro}. Applying $=_{\varphi}$ to the same system $x$ across different times yields a diachronic identity criterion for the system---see \eqref{eq:diachro} instead.

\bibliographystyle{ACM-Reference-Format}
\bibliography{sample-base}

\end{document}